\documentclass[conference]{IEEEtran}

\IEEEoverridecommandlockouts
\usepackage{cite}
\usepackage{amsmath,amssymb,amsfonts}
\usepackage{algorithmic}
\usepackage{graphicx}
\usepackage{textcomp}
\usepackage{xcolor}
\usepackage{soul}

\usepackage{dcolumn}
\usepackage{float}
\usepackage{caption}
\usepackage{tabularx}
\usepackage{subcaption}

\usepackage{stfloats}

\def\BibTeX{{\rm B\kern-.05em{\sc i\kern-.025em b}\kern-.08em
    T\kern-.1667em\lower.7ex\hbox{E}\kern-.125emX}}

\usepackage[english]{babel}

\usepackage{amsmath}
\usepackage{amssymb}
\usepackage{amsfonts}
\usepackage{mathtools}
\usepackage[binary-units=true]{siunitx}
\usepackage{caption}
\usepackage{subcaption}
\usepackage{graphicx}
\usepackage{xcolor}

\usepackage{booktabs}
\usepackage{tabularx}

\usepackage{tikz}
\usetikzlibrary{external,automata,trees,shadows,matrix,fit,decorations,positioning,shapes,plotmarks,patterns,arrows,calc,quotes,tikzmark,arrows.meta}

\makeatletter
\renewcommand{\vec}[1]{%
	\ifcat\relax\noexpand#1%
	\ensuremath{\boldsymbol{\lowercase{#1}}}%
	\else
	\ensuremath{\mathbf{\lowercase{#1}}}%
	\fi
}
\makeatother
\makeatother

\makeatletter

\newcommand{\argmin}[1]{\ensuremath{\underset{#1}{\text{argmin }}}}

\newcommand{\argmax}[1]{\ensuremath{\underset{#1}{\text{argmax }}}}
\renewcommand{\max}[1]{\ensuremath{\underset{#1}{\text{max }}}}

\usepackage{pifont}

\begin{document}

\title{A distributed neural network architecture for dynamic sensor selection with application to bandwidth-constrained body-sensor networks\\}

\author{Thomas Strypsteen and Alexander Bertrand, \IEEEmembership{Senior Member, IEEE}% <-this % stops a space
% \IEEEauthorblockA{\textit{KU Leuven, Department of Electrical Engineering (ESAT)} \\
% \textit{STADIUS Center for Dynamical Systems, Signal Processing and Data Analytics}\\
% Leuven, Belgium \\
% \{thomas.strypsteen, alexander.bertrand\}@esat.kuleuven.be}

\thanks{This project has received funding from the European Research Council (ERC) under the European Union’s Horizon 2020 research and innovation
programme (grant agreement No 802895). The authors also acknowledge the financial support of the FWO (Research Foundation Flanders) for project G.0A49.18N, and the Flemish Government under the “Onderzoeksprogramma Artifici\"ele Intelligentie (AI) Vlaanderen” programme.
\newline
T. Strypsteen and A. Bertrand are with KU Leuven, Department of Electrical Engineering (ESAT), STADIUS Center for Dynamical Systems, Signal Processing and Data Analytics and with Leuven.AI - KU Leuven institute for AI, Kasteelpark Arenberg 10, B-3001 Leuven, Belgium (e-mail: thomas.strypsteen@kuleuven.be, alexander.bertrand@kuleuven.be).}
}

\maketitle

\begin{abstract}
We propose a dynamic sensor selection approach for deep neural networks (DNNs), which is able to derive an optimal sensor subset selection for each specific input sample instead of a fixed selection for the entire dataset. This dynamic selection is jointly learned with the task model in an end-to-end way, using the Gumbel-Softmax trick to allow the discrete decisions to be learned through standard backpropagation. We then show how we can use this dynamic selection to increase the lifetime of a wireless sensor network (WSN) by imposing constraints on how often each node is allowed to transmit. We further improve performance by including a dynamic spatial filter that makes the task-DNN more robust against the fact that it now needs to be able to handle a multitude of possible node subsets. Finally, we explain how the selection of the optimal channels can be distributed across the different nodes in a WSN. We validate this method on a use case in the context of body-sensor networks, where we use real electroencephalography (EEG) sensor data to emulate an EEG sensor network. We analyze the resulting trade-offs between transmission load and task accuracy.
\end{abstract}


\section{Introduction}
\label{section: Section1}

Wireless sensor networks (WSNs) consist of a collection of networked wireless sensor nodes with local processing capabilities, which cooperate to solve a specific inference task \cite{akyildiz2002wireless, kandris2020applications, latre2011survey}. Due to technological advances in the miniaturization and energy-efficiency of sensors and microprocessors, such WSNs have become popular for long-term and wide-area monitoring in various domains such as acoustics, video surveillance, object tracking, and physiological sensing. In the latter case, such WSNs are also known as body area networks or body-sensor networks (BSNs), where physiological sensors at different locations on or in the body are wirelessly interconnected to share their data \cite{bertrand2015distributed,strypsteen2022bandwidth}.
\newline

Ensuring maximal battery lifetime is a crucial consideration in the design of these WSNs. The energy bottleneck will typically be found in the wireless transmission of the data between the sensors and/or a fusion center \cite{bertrand2015distributed}. This not only motivates energy-efficient hardware design, but also a shift in the algorithmic design of the models running on these sensor nodes. Instead of optimizing the model only for accuracy, the amount of data that needs to be transmitted when the model is used in the context of a WSN becomes an important design factor. 
\newline

In this paper, we focus on reducing this data transmission by teaching the nodes a policy where they only transmit their data when the contribution of this specific node towards the inference task would be very informative for the current input sample, while upholding a given computational constraint. Such a constraint could be, e.g., that each node can on average only transmit at most 50\% of its collected sensor data. 
To this end, we propose a \textit{dynamic channel selection} methodology. For each block of collected samples across the nodes of the WSN, a distributed dynamic channel selector computes an input-dependent, optimal subset of channels, represented by a binary mask across the channels. Inference is then performed on the masked input by a deep neural network (DNN) at a fusion center which collects the data transmitted by the sensors. The selector and the inference model are trained jointly in an end-to-end manner, with the discrete parameters involved in the selection process being made trainable through the Gumbel-Softmax trick \cite{jang2016categorical,maddison2016concrete}. How often each channel is selected is limited by a per-channel sparsity loss on the computed masks.
The usage of this dynamic channel selection means that the inference model will be presented with different channel subsets for different inputs, as if channels were randomly missing. We show that applying dynamic spatial filtering (DSF) \cite{banville2022robust} to the masked input to re-weight the channels helps the inference model become more robust against the missing of channels and improves performance.
\newline

To validate our proposed architecture, we focus on a specific use case in the area of brain-computer interfaces, where BSNs can be used to collect neural signals at different brain regions. One such an example is a BSN that monitors the brain via multi-channel electroencephalography (EEG) sensors, a so-called wireless EEG sensor network (WESN) \cite{bertrand2015distributed}. EEG is a widely used, noninvasive way to record electrical brain activity, measuring potentials on multiple locations on the scalp to yield multi-channel time signals. These signals contain useful information for a variety of tasks such as epileptic seizure detection \cite{ansari2019neonatal}, sleep stage analysis \cite{de2017complexity} and brain-computer interfacing (BCI) \cite{lawhern2018eegnet}. In a WESN, multiple lightweight mini-EEG devices record one or a few EEG channels from their respective scalp areas, process the data, and transmit it to other nodes or a central fusion center, rather than using a bulky cap as in traditional EEG recording methods. We note that, while our evaluation use case is focused on brain signals, our proposed methodology is generic and can be applied to other kinds of wireless sensor networks as well.
\\[10pt]
The main contributions of this paper are:
\begin{itemize}
    \item We propose an end-to-end learnable dynamic sensor or channel selection method that selects, for each window of a multi-channel input, an optimal subset of channels to use for inference, given a certain selection budget. This dynamic selection is learned jointly with the task DNN model and the Gumbel-Softmax trick is used to enable backpropagation for the discrete decisions involved.
    \item We demonstrate how this methodology can be used to reduce the transmission load in a wireless sensor network, thus increasing its battery lifetime. We do this by moving from centralized to distributed channel selection and enforcing per-node constraints to ensure a proper balancing of the transmission load. In addition, we present a use case where the method can improve the robustness of the classifier to noise bursts.
\end{itemize}
The paper is organized as follows. In section \ref{section: Section1} we go over previous work in static and dynamic feature selection. Section \ref{section: Section2} formally presents our problem statement and dynamic channel selection methodology. In section \ref{section: Section3} we provide an overview of the used dataset and how it was used to emulate a WESN environment and provide more details on the used model architecture and training strategy for this specific experiment. Our experimental results are then presented in section \ref{section: Section4} and we end with some conclusions in section \ref{section: Section5}.

\textbf{Note on terminology:} Throughout this paper, we will always use the term `channel selection' to refer to a selection of channels from a multi-channel input signal. Sensor selection or node selection could be viewed as a special case of channel selection. In the case of single-channel sensors, sensor or channel selection refer to the same thing. However, in the case of multi-channel sensors, sensor selection  refers to the problem of selecting pre-defined \textit{groups} of channels rather than individual channels, where each group corresponds to a sensor. For the sake of an easy exposition, but without loss of generality, we will assume single-channel sensors throughout this paper. Sometimes, we will refer to sensors as `nodes' for consistency in terminology with the WSN literature.

\section{Related work}
\label{section: Section1}

% \subsection{EEG channel selection}
\subsection{Static feature selection}

\label{section: Section1a}

The goal of feature selection is to find an optimal subset of an available set of features that maximizes the performance of a classification or regression model on a given task. A host of literature exists that solves this problem in a \textit{static} way, i.e., the optimal subset is determined for a certain dataset as a whole and the same selection is then applied to all input samples. Filter-based approaches rank the available features by a criterion like mutual information (MI) with the target labels and select the $K$ highest scoring features \cite{lan2006salient}. Wrapper-based approaches use methods like greedy backward selection to efficiently explore the space of possible feature subsets, train the model on these candidate subsets and finally select the one that performs the best \cite{narayanan2019analysis}. Embedded approaches jointly learn the subset and the task model in an end-to-end way, by performing $L_1$ regularization on the input weights \cite{scardapane2017group} or learning the discrete parameters of the feature selection using continuous relaxations \cite{abid2019concrete,strypsteen2021end}. In this paper, we employ this approach of continuous relaxations to perform \textit{dynamic} channel selection instead.
\subsection{Dynamic feature selection}
\label{section: Section1b}

In \textit{dynamic}, or \textit{instance-wise} feature selection, the aim is to find an optimal subset of features for each individual input sample. One area where this approach has been highly relevant is field of explainable machine learning, where the goal is to indicate which features contributed most to the model output. For instance, L2X \cite{chen2018learning} trains an explainer model that maximizes the mutual information between the feature subset of size $K$ of a given sample and the class distribution yielded by a trained task model. This line of work however, is mainly interested in finding the most relevant features for an already trained model, not in optimizing the performance of the model on reduced feature sets.
\newline

Another relevant area is the field of active feature acquisition \cite{li2021active,covert2023learning}. In this setting, obtaining features is associated with a certain cost. The goal is then to obtain maximal model performance with a minimal amount of features, without being able access all the features of a given input sample from the start. This typically results in an iterative procedure where, based on the current feature subset, the optimal feature to extend the set with is estimated, until sufficient confidence in the model prediction is reached or the budget is saturated. In our WSN setting in contrast, we do have access to all the features to perform subset selection, but we are not allowed to centralize all of them by transmitting them over the wireless link. We also aim to avoid iterative procedures that require multiple communication rounds between the sensors and the fusion center as these are prone to latency issues in real-time situations.
\newline

The most similar approach to ours in terms of methods is taken by Verelst et al. in the field of computer vision \cite{verelst2020dynamic}. The aim of their work is to decrease the computation time and energy of a CNN by learning input-dependent binary masks that are applied to the feature maps of an image at each layer. That layer then only performs convolutions on the pixels that are not masked out. A sparsity loss on these masks then forces the network to adhere to a certain computational budget, with backpropagation for the discrete masks being enabled through the Gumbel-Softmax trick. We will employ a similar strategy to learn binary masks for our dynamic selection, albeit in a distributed architecture and with the goal of reducing the data transmission over the wireless links, as detailed in the next section.

\begin{figure*}
    \centering
    \includegraphics[width=0.95\textwidth]{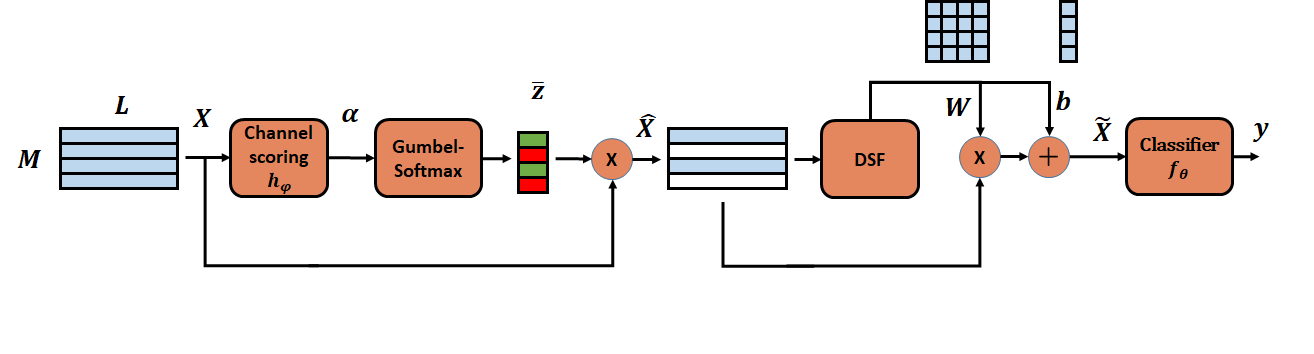}
    \caption{Overview of the dynamic channel selection. An $L$-sample window of an M-channel signal is passed to a channel scoring module, yielding unnormalized channel scores $\vec{\alpha}$. These are then converted in discrete selections by the Gumbel-Softmax module and applied to the input as a binary mask $\bar{\vec{z}}$, dropping a number of channels. This masked input $\hat{X}$ is then fed to the DSF module which re-weights the received channels with an attention mechanism, computing a weight matrix $W$ and bias $b$ that are multiplied with and added to the masked input: $\tilde{X} = W\hat{X} + b$ \cite{banville2022robust}. Finally, the original classifier is then applied to the resulting signal to obtain a prediction $y$. The entire pipeline is jointly learned in an end-to-end manner.}
    \label{fig: overview}
\end{figure*}

\section{Proposed Method}
\label{section: Section2}

In this section, we describe our dynamic channel selection methodology. Without loss of generality, we assume a classification task, although all methods can be easily extended to a regression task. For the sake of an easy exposition, we will initially assume a centralized architecture where all the channels are available to make a decision on the selection. In a WSN context, this implies that the channel selection is performed \textit{after} transmitting all the channels to the fusion center, in which case there are no bandwidth savings. Nevertheless, this setting is still relevant to make the network robust against non-stationary noise influences and/or to reduce the computational complexity at the fusion center. Later on, in Section \ref{section: Section2e}, we will explain how the channel selection can be performed at the level of the sensor nodes, such that non-selected channels do not have to be transmitted at all. 

\subsection{Problem statement}
\label{section: Section2a}

Let $\mathcal{D} = \{(X^{(1)},y^{(1)}), (X^{(2)},y^{(2)}), \ldots , (X^{(N)},y^{(N)})\}$ be a dataset of $N$ samples of a multi-channel signal $X^{(i)}$ with class labels $y^{(i)}$. Each $X^{(i)} \in {\rm I\!R}^{M \times L}$ contains $M$ channels and a window of $L$ consecutive time steps. We are also given a DNN model $f_\theta$ that is used to perform inference on these samples.
Our goal is to learn a dynamic selection function that, for each separate input sample, determines this sample's optimal subset of channels to be presented to the inference model, while adhering to certain budget constraints on the amount of channels we are allowed to use on average. This selection of channels is based on a score vector $\vec{\alpha} \in \mathbb{R}^{M}$ that is computed by a channel scoring function $h_\varphi(X)$ for each input $X$. To go from a continuous score to a discrete selection in a way that still allows for end-to-end learning through backpropagation, we make use of a Gumbel-Softmax module $G$, which converts the score vector $\vec{\alpha}$ to a binary mask $\bar{\vec{z}} \in \{0,1\}^{M}$ to be applied to the input. Formally, this means learning parameters $\theta$ of the task model $f_{\theta}$ and the parameters $\varphi$ of a selection model $s_\varphi = G \circ h_\varphi: \mathbb{R}^{M \times L} \mapsto \{0,1\}^{M \times 1}; X \mapsto \bar{\vec{z}}$ such that

\begin{equation}
\label{eq: loss}
\begin{split}
    \varphi^*, \theta^* = & \argmin{\varphi, \theta} \mathcal{L}_{CE}(f_{\theta}(X \odot \bar{\vec{z}}\mathbf{1}_L^\top),y) + \lambda \mathcal{L}_{S}(\bar{\vec{z}}) \\
     = & \argmin{\varphi, \theta} \mathcal{L}_{CE}(f_{\theta}(X \odot s_{\varphi}(X)\mathbf{1}_L^\top),y) + \lambda \mathcal{L}_{S}(s_{\varphi}(X))
\end{split}
\end{equation}
with $\mathcal{L}_{CE}(p,y)$ the cross-entropy loss between the predicted label $p$ and the ground truth $y$, $\odot$ an element-wise product, $\mathbf{1}_L^\top$ the row vector of dimension $L$ containing only ones, $\mathcal{L}_S$ a cost function that enforces sparsity in the learned masks and $\lambda$ a hyperparameter to balance the two losses. A schematic overview of our method is presented in Fig. \ref{fig: overview}. We will now delve deeper into the design of each of the modules involved.

\subsection{Learning discrete decisions with Gumbel-Softmax}
\label{section: Section2b}

To enable the network to learn discrete decisions while still keeping the entire network end-to-end learnable we make use of the Gumbel-Softmax trick \cite{maddison2016concrete,jang2016categorical}. Take a discrete random variable, drawn from a categorical distribution with $K$ classes and class probabilities $\pi_1,...\pi_K$, represented as a one-hot vector $\mathbf{\bar{y}} \in \{0,1\}^{K}$, with the index of the one indicating the class $\mathbf{\bar{y}}$ belongs to. Discrete samples from this distribution can then be drawn with the Gumbel-Max trick:
\begin{equation}
    \mathbf{\bar{y}}=\text{one\_hot}(\argmax{k} (\log \pi_k + g_k))
\end{equation}
with $g_k$ independent and identically distributed (i.i.d.) samples from the Gumbel distribution \cite{gumbel1948statistical} and $\text{one\_hot}(i)$ the operator that generates a one-hot $K\times1$ vector where the one is placed at position $i$. The Gumbel-Softmax is then a continuous, differentiable relaxation of this discrete sampling procedure, approximating the discrete one-hot vectors $\mathbf{\bar{y}}$ with continuous vectors $\mathbf{y}$ whose elements sum to one instead by replacing the argmax with a softmax. For the k-th element $y_k$, this results in:
\begin{equation}
\label{eq: sampling}
    y_k = \frac{\exp((\log \ \pi_{k} + g_{k})/\tau)}{ \sum_{j=1}^{K} \exp((\log \ \pi_{j} + g_{j})/\tau)}    
\end{equation}
with $\tau$ the temperature of this continuous relaxation. Lowering the temperature causes the softmax to more closely resemble an argmax, thus causing the continuous $\mathbf{y}$ to be a closer approximation of the discrete $\mathbf{\bar{y}}$. It will however, also cause the relaxation to become less smooth and increase the variance of the gradients.
\newline

\begin{figure*}
    \centering
    \includegraphics[trim={1cm 0cm 1cm 0cm },height=0.22\textheight]{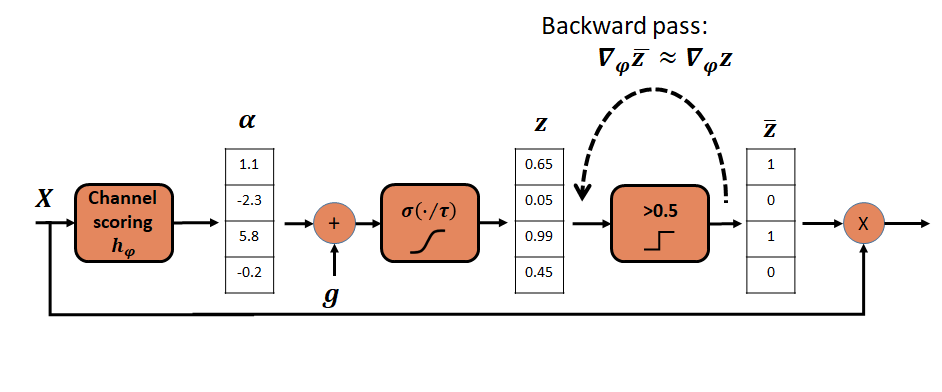}
    \caption{Illustration of the Gumbel-Softmax trick. During training, hard decisions are sampled by perturbing the channel scores $\vec{\alpha}$ with Gumbel noise, passing this through a sigmoid to obtain soft probabilities and applying a thresholding operator. Backpropagation through the thresholding operation is enabled by using a Straigth-Through estimator, treating the threshold in the backward pass as an identity function, i.e. $\frac{\partial\bar{\vec{z}}}{\partial\vec{z}} \approx 1$ and thus $\nabla_\varphi \bar{\vec{z}} \approx \nabla_\varphi \vec{z}$  }
    \label{fig: gumbel}
\end{figure*}

Our goal is to model a learnable, binary random variable $\bar{z}_m$ for each channel $m$, which is 1 when the channel is selected and 0 otherwise. In the case of such a binary random variable $\bar{z}_m$, with $P(\bar{z}_m=1)=\pi_1$, it can be shown \cite{verelst2020dynamic} that by setting $K=2$ in Eq. \ref{eq: sampling}, the Gumbel-Softmax trick can be simplified to
\begin{equation}
\label{eq: binary}
\begin{split}
    y_1 & =\sigma\left(\frac{\log \pi_1 + g_1 - g_2}{\tau}\right) \\
    y_2 & =1-y_1
\end{split}
\end{equation}
with $\sigma(\cdot)$ the sigmoid function. A continuous relaxation $z_m$ of the binary random variable $\bar{z}_m$ can then be obtained by taking $z_m = y_1$ We can use this binary Gumbel-Softmax trick to transform unnormalized, learnable channel scores $\vec{\alpha} \in {\rm I\!R}^{M}$ yielded by a network $h_\varphi(X)$ into continuous, differentiable approximations $\vec{z} = [z_1, ..., z_M]^\top$ of the discrete $\bar{\vec{z}} \in \{0,1\}^{M}$. There are a number of ways this continuous relaxation can be used to obtain approximating gradients for the discrete $\bar{\vec{z}}$, but we will follow the Straight-Through estimator approach \cite{verelst2020dynamic,jang2016categorical}. This means that we will sample discrete decisions from our binary distribution in the forward pass:
\begin{equation}
    \begin{split}
    \bar{z}_m = & \left \lfloor  \sigma\left(\frac{\alpha_m + g_1 - g_2}{\tau}\right) \right \rceil \\
    = &
    \begin{cases}
      1, & \text{if}\ z_m = \sigma\left(\frac{\alpha_m + g_1 - g_2}{\tau}\right) > 0.5 \\
      0, & \text{otherwise}
    \end{cases} \\
    \end{split}
\end{equation}
with $\lfloor\cdot\rceil$ the rounding operator, resulting in a binary distribution where $P(\bar{z}_m=1)=\sigma(\alpha_m)$ (replacing $\pi_1$ in Eq. \ref{eq: binary}). To enable backpropagation through the discrete rounding operator, we use gradients from the continuous relaxation in the backward pass, which implies the approximation
\begin{equation}
    \nabla_\varphi {\bar{\vec{z}}} \approx \nabla_\varphi \vec{z}.
\end{equation}
This scheme allows for hard decisions to be used during training and learned through end-to-end backpropagation. This process is schematically illustrated in Fig. \ref{fig: gumbel}. At inference time, Gumbel noise is no longer added to the score vector, resulting in the network no longer sampling from binary distributions, but behaving in a deterministic manner instead, i.e. $\bar{z}_m=1$ if $\sigma(\alpha_m)>0.5$.

% One way is annealing the temperature during training, causing $\vec{z}$ to become more and more discrete as training progresses and thus $\nabla_\varphi \vec{z}$ to converge to $\nabla_\varphi \bar{\vec{z}}$. Th

\subsection{Enforcing sparsity}
\label{section: Section2c}

We assume each channel is measured on a different node of a wireless sensor network, whose nodes are able to communicate with each other over bandwidth-constrained links. To reduce the communication load of these nodes, we want each node $m$ to only transmit their data (i.e. yield a 1 in their binary mask) for a maximal target percentage $ T \in [0,1] $ of the input samples. We thus want the expected value of each element of the binary mask over the distribution of our input samples $X$ to be below this target $T$:
\begin{equation}
\label{eq: constraints}
\begin{split}
    & \mathbb{E}_{X}[\bar{z}_m] \leq T \\
    & m=1,..,M
\end{split}
\end{equation}

These per-node constraints ensure that the masks are not only sparse, but balanced across the different nodes as well, meaning that there is no single node that is transmitting significantly more than the others. Secondly, by applying the constraints to the expected value of the masks instead of on each separate masks, we allow for a variable amount of nodes to be used for different input samples. To enforce these constraints, we use a mini-max optimization in which we impose a per-node sparsity loss on the decisions made during training and aggregate these by penalizing the node that currently violates this constraint the most:
\begin{equation}
\label{eq: sparsity}
\begin{split}
    \mathcal{L}_{S,m} & = \max{} \left( \frac{1}{B}\sum_{b=1}^B \sigma\left(\frac{\alpha_m^{(b)}}{\tau_0}\right) - T, 0 \right)^2 \\
    \mathcal{L}_{S} & = \max{m} \mathcal{L}_{S,m}
\end{split}
\end{equation}
with $B$ the batch size, $\alpha_m^{(b)}$ the score for node $m$ for the b'th input sample of the batch and $\tau_0$ a temperature constant we set at $0.1$. This sparsity loss replaces the expected value in the constraints of Eq. \ref{eq: constraints} with a batch average and the discrete node decisions $\bar{z}_m$ with the continuous approximation $\sigma\left(\frac{\alpha_m^{(b)}}{\tau_0}\right)$. The fact that this approximation is computed through a sigmoid with a low temperature, without the addition of Gumbel noise means we more closely approximate the behaviour of the selection layer at inference time than we would if we directly penalized the hard decisions $\bar{\vec{z}}$. This is important to ensure that if the sparsity constraints are met at training time, they will also be met at inference time. For instance, if the network ensures that for all $X$, $\sigma(\alpha_m)=0.51$, then due to the addition of Gumbel noise in the computation of $\bar{z}_m$, the network will sample $\bar{z}_m=1$ and $\bar{z}_m=0$ an about equal amount of times during training. At inference time however, when no noise is added, the network will always yield $\bar{z}_m=1$ since $\sigma(\alpha_m)>0.5$, surely violating the constraints. Also important to note is that using Eq. \ref{eq: sparsity} requires training with a large enough batch size such that the batch average of Eq. \ref{eq: sparsity} is a meaningful estimate of the expected value used in Eq. \ref{eq: constraints}.

%The reason whe do not use zbar is that due to the addition of noise in its computation, the network could cheat by putting al probabilities sigmoid(alpha/temp) to 0.51, easily satisfying the constraints due due to the noise ensuring we will jump around this value at train time, but having 100% usage at test time

\subsection{Dealing with different channel subsets}
\label{section: Section2d}
Performing dynamic channel selection means the classification network $f_\theta$ will see a different subset of active channels depending on the current input sample. Stated in another way, our network needs to be able to deal with missing inputs. This can cause problems in the learning of the network weights, as the network needs to be able to extract relevant information when a channel is selected, but not cause interference when the channel is not selected and the corresponding input only contains zeros. Ideally, we would employ a number of separate classification networks, each optimized for a specific channel subset. In practice however, this would require training and storage of $2^M$ networks, which quickly becomes infeasible. Thus, the question arises how we can make a single network be able to cope as efficiently as possible when multiple input sets are possible. We tackled this issue by extending our network with the Dynamic Spatial Filtering (DSF) proposed by Banville et al. \cite{banville2022robust}. The idea of DSF is to re-weight the $M$ input channels using an attention layer. In this setting, new (virtual) channels are formed by applying a spatial filter to all input channels, i.e., making linear combinations of the channels, with the weights being computed from the spatial covariance matrix of the current input window. This re-weighting decreases the impact of missing channels on the network activations and has been shown to make a network more robust against noisy or missing channels.

\subsection{From centralized to distributed}
\label{section: Section2e}

\begin{figure*}[]
\centering
\begin{subfigure}{0.33\textwidth}
   \captionsetup{font={small}}

  \centering
    \includegraphics[trim={0cm 0cm 0cm 0cm},height=0.22\textheight]{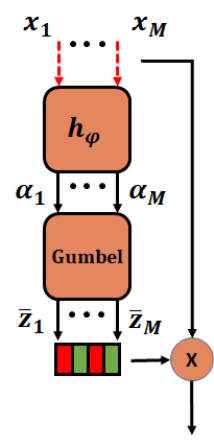}  \caption{Centralized}
  \label{fig:sub1}
\end{subfigure}%
\begin{subfigure}{0.33\textwidth}
   \captionsetup{font={small}}

  \centering
    \includegraphics[trim={0cm 0cm 0cm 0cm},height=0.22\textheight]{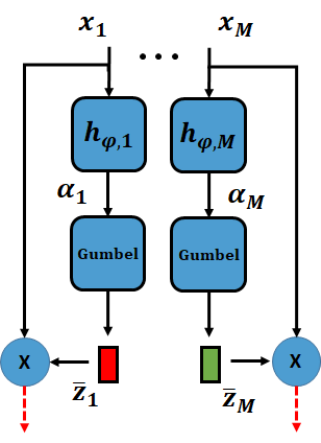}  \caption{Distributed}
  \label{fig:sub1}
  \end{subfigure}
  \begin{subfigure}{0.33\textwidth}
   \captionsetup{font={small}}

  \centering
    \includegraphics[trim={-1.1cm 0.0cm 0cm 0cm},height=0.22\textheight]{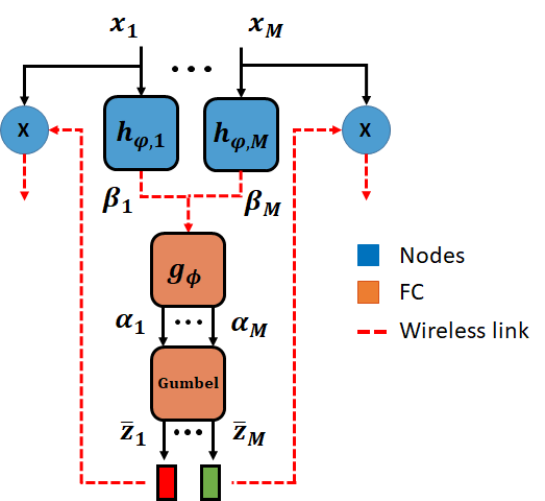}  \caption{Distributed-Feedback}
  \label{fig:sub1}
  \end{subfigure}
\caption{Different settings for the channel scoring module in a sensor network. Blue blocks indicate modules on the sensor nodes, orange modules on the fusion center and red dotted lines communications between the two. (a) The centralized upper bound employs all joint information of all channels to compute a score, but would require all data to be centralized. (b) The distributed setting does not allow for communication between the nodes (via the fusion center) and makes each score only dependent on the local data. (c) The distributed-feedback setting allows for a small amount of communication between the nodes to make a better, joint decision.}
\label{fig: distributed}
\end{figure*}

The channel scoring function $h_\varphi$ in Eq. \ref{eq: loss} currently still uses all M input channels to make a decision. However, an important aspect to be taken into account is the distributed nature of WSN platforms, where different channels are recorded on different physical devices. In this setting, we want to reduce the transmission load of these devices by only selecting and thus transmitting the signal of a node when its information is relevant for the current sample. However this will only actually be beneficial when we are able to perform the selection \textit{without centralizing the data of the different sensors}. We will consider three different cases corresponding to different constraints on our dynamic channel scoring function $h_\varphi(X)$:
\begin{itemize}
    \item \textit{Centralized}: The selection is derived from the joint information of all channels, i.e., $\vec{\alpha} = h_\varphi(X)$. This setting serves as a theoretical upper bound for the following two practical settings.
    \item \textit{Distributed}: Each node has to decide whether to transmit solely on its own data, i.e., $\alpha_m = h_{\varphi,m}(\vec{x}_m)$ where $x_m$ denotes the $m$-th row of $X$.
    \item \textit{Distributed-Feedback}: Each node computes a short vector $\vec{\beta}_m=h_{\varphi,m}(\vec{x}_m) \in \mathbb{R}^C$ with $C << L$, that is transmitted to the fusion center. At the fusion center, the $\vec{\beta}_m$ of all nodes are combined into the stacked vector $\vec{\beta}$  to determine a final scoring vector $\vec{\alpha} = g_\phi(\vec{\beta})$. The discrete selection $\bar{\vec{z}}$ resulting from this scoring vector is then returned to the nodes to inform them which of them should transmit. The size of these vectors $\vec{\beta}_m$ should be small compared to the length $L$ of the window to be transmitted to minimize the overhead cost of the selection. In order to reduce the trainable parameters, one can decide to make the different $h_{\varphi,m}$ models copies of each other, with shared weights for all layers, except the final layer having its own set of parameters for each channel $m$.
\end{itemize}
These three settings are illustrated in Fig. \ref{fig: distributed}.

\subsection{Training strategy}
\label{section: Section 2f}

Successfully training a model with masking units typically hinges on a good initialization. Since a sparsity loss is much easier to minimize than the training objective - by simply driving the weights of the binary masks to zero - the network can quickly collapse into a state where barely any units are executed \cite{verelst2020dynamic,figurnov2017spatially}. Once this has happened, it is very hard for the network to learn task-relevant information that could pull it out of this state. To avoid this, we adopt a step-wise training strategy that learns one module at a time. 
\begin{enumerate}
    \item Initialize the weights of the classifier $f_\theta$ with the weights of the original M-channel network trained without any dynamic selection.
    \item Add the \textit{centralized} dynamic selection layer and train it while fine-tuning (i.e., training at a lower learning rate) the classifier. 
    \item Add the DSF module and train it while fine-tuning the the dynamic selection and classifier.
    \item Transform the centralized dynamic selection layer in a distributed dynamic selection layer and fine-tune the whole model (see below).
\end{enumerate}

To go from a centralized to a distributed architecture, we employ a 2-step transfer learning approach. First, we employ the centralized channel scoring function $h_\varphi$ as a teacher model and try to ensure that the outputs $\vec{\alpha}_{distr}$ of the student model - the distributed channel scoring function - match the outputs $\vec{\alpha}_{centr}$ of the teacher by minimizing the following loss:
\begin{equation}
    \mathcal{L}(h_{\varphi,distr}) = \mathcal{L}_{BCE}( \sigma(\vec{\alpha}_{distr}), \lfloor\sigma(\vec{\alpha}_{centr})\rceil)
\end{equation}
with $\mathcal{L}_{BCE}$ the binary cross-entropy loss. By minimizing this loss, we do not necessarily ensure that the channel scores $\vec{\alpha}$ are exactly alike, but rather that the discrete outputs at inference time will be similar, which is what we ultimately want. In the final step, we use the newly learned distributed channel selection layer, initialize the DSF module and the classifier with the corresponding weights of the centralized model and fine-tune the whole network in an end-to-end fashion.

% \begin{figure}[]
% \centering
% \begin{subfigure}{0.16\textwidth}
%   \centering
%     \includegraphics[trim={0.5cm 0.0cm 1.25cm 0.0cm},width=\textwidth]{Distributed1.png}  \caption{Centralized}
%   \label{fig:sub1}
% \end{subfigure}%
% \begin{subfigure}{0.16\textwidth}
%   \centering
%     \includegraphics[trim={0.5cm 0.0cm 1.25cm 0.0cm},width=\textwidth]{Distributed2.png}  \caption{Fully Distributed}
%   \label{fig:sub1}
%   \end{subfigure}
%   \begin{subfigure}{0.16\textwidth}
%   \centering
%     \includegraphics[trim={1.25cm 0.0cm 0.5cm 0.0cm},width=\textwidth]{Distributed3.png}  \caption{Distributed}
%   \label{fig:sub1}
%   \end{subfigure}

\section{Application to Wireless EEG Sensor Networks}
\label{section: Section3}

\subsection{Data set}
\label{section: Section3a}

In the field of BCI, the motor execution paradigm is used to decode body movement from the corresponding neural signals in the motorsensory areas of the brain. The High Gamma Dataset \cite{schirrmeister2017deep} contains EEG data from about 1000 trials of executed movements for each of the 14 subjects, as well as a separate test set of about 180 trials per subject, all following a visual cue. The dataset includes 4 classes of movements: left hand, right hand, feet, and rest. As in \cite{schirrmeister2017deep} we only use the 44 channels that cover the motor cortex, which are preprocessed by resampling at 250 Hz, highpass filtering above 4 Hz, standardizing the per-channel mean and variance to 0 and 1 respectively, and extracting a window of 4.5 seconds for each trial. This pre-processing is adopted from \cite{schirrmeister2017deep} and described in full detail there.

\subsection{WESN node emulation and selection}
\label{section: Section3b}

In mini-EEG devices, we cannot measure the potential between a given electrode and a distant reference (e.g. the mastoid or Cz electrode) , as we would in traditional EEG caps \cite{narayanan2019analysis}. Instead, we can only record the local potential between two nearby electrodes belonging to the same sensor device. To emulate this setting using a standard cap-EEG recording, we follow the method proposed in \cite{narayanan2019analysis}, which considers each pair of electrodes within a certain maximum distance as a candidate electrode pair or node. By subtracting one channel from the other, we can remove the common far-distance reference and obtain a signal that emulates the local potential of the node. Applying this method with a distance threshold of 3 cm to our dataset, we obtain a set of 286 candidate electrode pairs or nodes, which have an average inter-electrode distance of 1.98 cm and a standard deviation of 0.59 cm.
\newline

% In traditional EEG caps, a channel is usually measured as the potential between an electrode at a given location and a common reference, typically the mastoid or Cz electrode. However, in the case of mini-EEG devices, we can only measure a local potential between two proximate electrodes belonging to the same device. To emulate this setting based on a standard cap-EEG recording, we follow the approach of \cite{narayanan2019analysis}. In this setting, each pair of electrodes within a preset maximum inter-electrode distance from each other is a candidate electrode pair or node we could measure. The signal this node records is then  emulated by subtracting one of the channels from the other, thus removing the common (far-distance) reference in the process. We applied this method with a distance threshold of 3 cm to our dataset, converting the 44 channels in 286 candidate electrode pairs or nodes. The resulting set of nodes had an average inter-electrode distance of 1.98 cm with a standard deviation of 0.59 cm.

Given that our WESN will consist of a limited number of mini-EEG devices, we first need to select the $M$ most informative sensor nodes from the pool of 286 candidate nodes. To achieve this, we adopt the static channel selection method described in \cite{strypsteen2021end}, which enables us to learn the $M$ optimal nodes for a given task and neural network by jointly training the network and a  selection layer. Note that this is a fixed selection for the entire data set, not for each sample separately. We train this selection layer, along with the centralized network ($f_\theta$ in Fig. \ref{fig: overview} we will use for classification (see Section \ref{section: Section3c}), using data from all subjects in the dataset, which results in a subject-independent set of $M$ mini-EEG nodes that are best suited for solving the motor execution task. We do this for 3 different values of $M$, corresponding to a small WESN ($M=4$ nodes), a medium-size WESN ($M=8$ nodes), and a high-density WESN ($M=16$ nodes).
\newline
% Since we are only able to use a limited number of mini-EEG devices, we will first perform a channel/node selection step to select the most relevant sensor nodes from the pool of 286 candidate nodes. To this end, we employ the regularized Gumbel-Softmax method described in \cite{strypsteen2021end}. This method allows us to learn the $M$ optimal nodes for a given task and neural network by training said network jointly with a special selection layer that is able to learn the discrete variables involved in feature selection through simple backpropagation. The value of $M$ will also be varied throughout our experiments. We jointly train this selection layer of size $M$ with the centralized MSFBCNN architecture using the data from all subjects in the data set, resulting in a subject-independent set of $M$ mini-EEG nodes that are optimally placed to solve the motor execution task. The $M$ selected nodes are then used to design a distributed version of the MSFBCNN network as explained next.

\subsection{Model architecture}
\label{section: Section3c}

As mentioned above, the neural network architecture we employ for classification ($f_\theta$ in Fig. \ref{fig: overview}) is the MSFBCNN proposed in \cite{wu2019parallel}, which was designed specifically for a motor execution task. Inspired by the Filterbank-CSP approach of \cite{ang2008filter}, this model computes log-power features by applying a number of temporal filters in parallel, aggregating these with spatial filters, and then applying squaring and average pooling over time. These features are then classified by a single linear layer. While the details of this network are not relevant for this study, we provide a summary of this network in table format in Appendix A for completeness.
\newline

For our channel scoring module $h_\varphi$ in the centralized setting, we will employ the same architecture, with the final linear layer being adapted to output a vector of dimension $M \times 1$. In the two distributed settings, the different node scoring models $h_{\varphi,m}$ are copies of each other, with shared weights for all layers, except the final layer having its own set of parameters for each node/channel $m$. The network architecture of these $h_{\varphi,m}$ is simply a single-input version of the M-input network define by $f_\theta$, but where the last fully-connected layer outputs the scalar $\alpha_m$ in the distributed setting and the node summary $\vec{\beta}_m \in \mathbb{R}^{C \times 1}$ in the distributed-feedback setting. The dimension of these node summaries $\vec{\beta}_m$ is chosen to be $C=10$, ensuring the overhead of its transmission is negligible compared to the transmission of the full window of $L=1125$ time samples.The module $g_\phi(\vec{\beta})$ aggregating the node summaries in the fusion center is a simple 2-layer multilayer perceptron (MLP) with a hidden dimension of 50 and ReLU nonlinearity. The DSF module also consists of a 2-layer MLP with hidden dimension 50 and ReLU nonlinearity, which is applied to the vectorized sample covariance matrix $\frac{1}{L} \hat{X} \hat{X}^\top$  of the masked input sample and which produces a weight matrix $W \in \mathbb{R}^{M \times M}$ and a bias $b \in \mathbb{R}^{M \times 1}$ which are used to compute a re-weighted output $\tilde{X} = W\hat{X} + b$.
\newline

Finally, for training, we follow the procedure described in section \ref{section: Section 2f}, using the Adam optimizer \cite{kingma2014adam} with a learning rate of $10^{-3}$ when a module is trained for the first time and $10^{-4}$ when it is fine-tuned during subsequent steps. A batchsize of 64 is employed and training lasts for 100 epochs with early stopping activated when the validation loss does not decrease for 10 epochs. The hyperparameter $\lambda$, controlling the penalization of the sparsity loss was set to 10 for this application and a fixed temperature $\tau=1$ was used for the Gumbel-Softmax module.

\section{Experimental Results}
\label{section: Section4}

\subsection{Centralized versus distributed performance}
\label{section: Section4a}

We first analyze the rate-accuracy trade-off obtained by our proposed dynamic channel selection method and investigate the impact of going from a theoretical, centralized approach to a practical, distributed one. To do this, we train our model for a given target rate $T$, indicating the maximal percentage of input samples for which the data of each node should be used and transmitted. Since the lifetime of the WSN as a whole will ultimately be determined by the node with the highest transmission rate, we will always report the \textit{maximal} rate $\mathcal{R}_{max}$ among the nodes, rather than the \textit{average}. As a proof-of-concept benchmark, we will employ a naive system that for each sample and for each node randomly determines whether the data should be transmitted, doing so with a probability equal to the relative target rate. This allows us to investigate whether the dynamic selection is truly able to make intelligent decisions that go beyond simply making sure the constraints are met. We train our model for a range of target rates and for networks consisting of 4, 8 and 16 nodes, each time averaging the results over 5 runs.
\newline

\begin{figure*}[htbp]
\centering
\begin{subfigure}{.32\textwidth}
  \centering
    \includegraphics[trim={1cm 0 0cm 0},width=\textwidth]{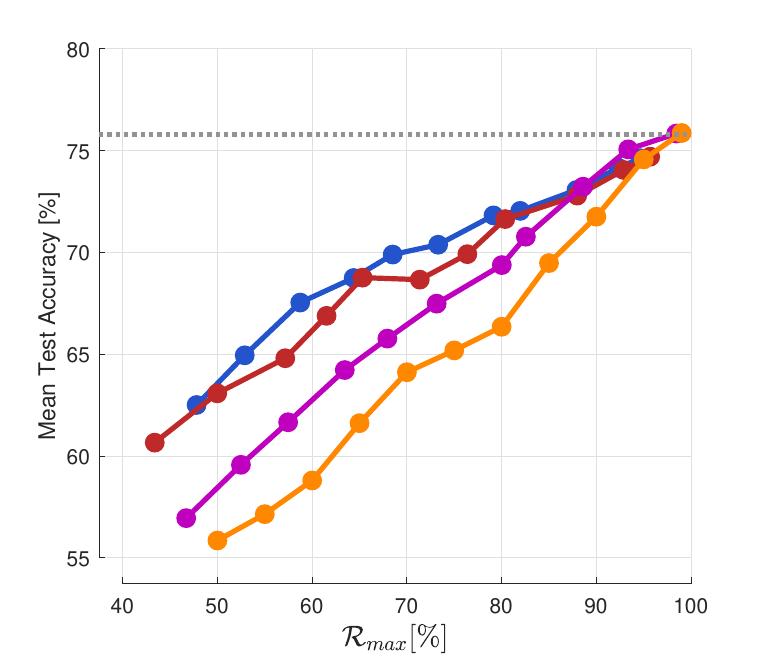}  \caption{$M=4$ nodes}
  \label{fig:sub1}
\end{subfigure}%
\begin{subfigure}{.32\textwidth}
  \centering
    \includegraphics[trim={1cm 0 0cm 0},width=\textwidth]{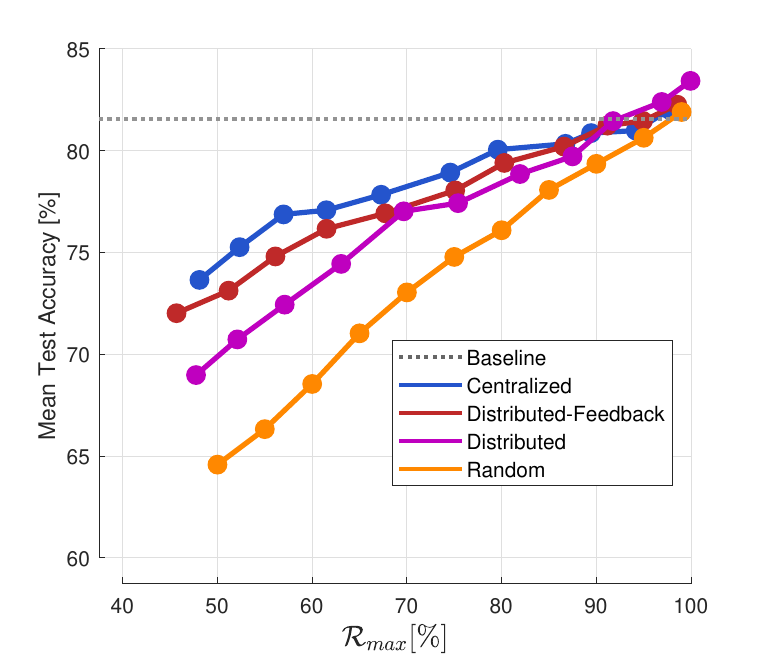}  \caption{$M=8$ nodes}
  \label{fig:sub1}
  \end{subfigure}
  \begin{subfigure}{.32\textwidth}
  \centering
    \includegraphics[trim={1cm 0 0cm 0},width=\textwidth]{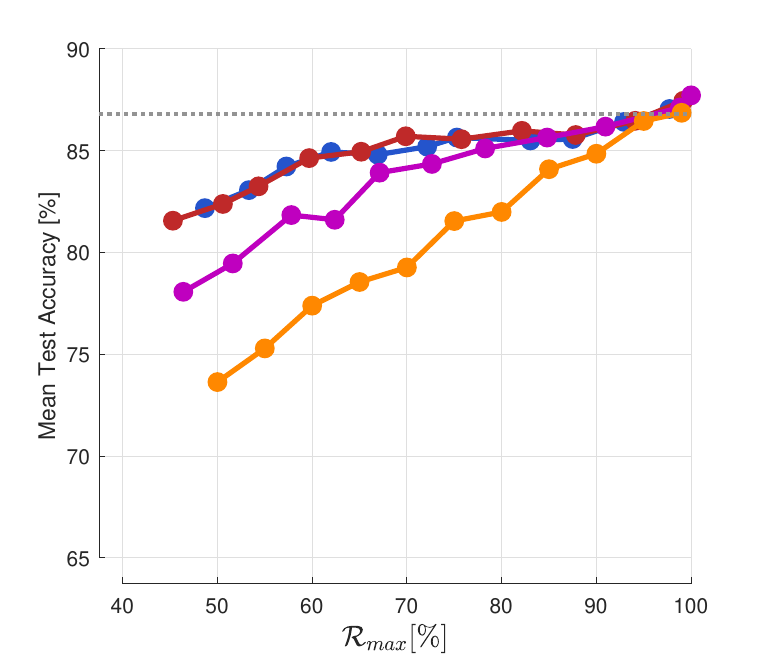}  \caption{$M=16$ nodes}
  \label{fig:sub1}
  \end{subfigure}
\caption{Rate-accuracy trade-off for the proposed dynamic channel selection method for networks of 4,8 and 16 nodes. Mean test accuracies are plotted against the percentage of samples for which the critical node of the network needs to transmit, i.e. the node with the highest percentage of transmission. Each data point is an average of 5 runs for a given maximal target rate $T$ (see Eqs. \ref{eq: constraints} and \ref{eq: sparsity}). Baseline performance indicates accuracy without dynamic selection involved, i.e. each node transmits at a rate $\mathcal{R}$ of 100\%. Dynamic selection consistently outperforms random channel selection. While a gap exists between the distributed implementation and the centralized upper bound, sharing a small amount of information between the nodes in the distributed-feedback setting largely overcomes this and performs about as well as the centralized upper bound.}
\label{fig:centralized-distributed}
\end{figure*}

Fig. \ref{fig:centralized-distributed} shows the resulting rate-accuracy tradeoffs. Firstly, we can observe that the dynamic selection indeed consistently outperforms the random selection, with the gap widening as the target rate decreases. Secondly, while the distributed network shows a small performance gap with the centralized upper bound, allowing for even a small amount of communication between the nodes by employing the distributed-feedback setting compensates for this as it performs very similarly to the centralized setting. When comparing the networks with a different amount of nodes, it can be observed that the more nodes are being used, the smaller the relative performance losses are when moving from the starting rate of 100\% to a rate of 50\% (for instance, the 16-node network only loses 5\% accuracy, while the 4-node network loses 10\%). Since there will be a higher amount of redundancy between the 16 nodes than between the 4 nodes, it makes sense that dropping channels in the former has less of an impact on the accuracy than in the latter. 

\subsection{Impact of dynamic spatial filtering}
\label{section: Section4b}

\begin{figure*}[htbp]
\centering
\begin{subfigure}{.32\textwidth}
  \centering
    \includegraphics[trim={1cm 0 0cm 0},width=\textwidth]{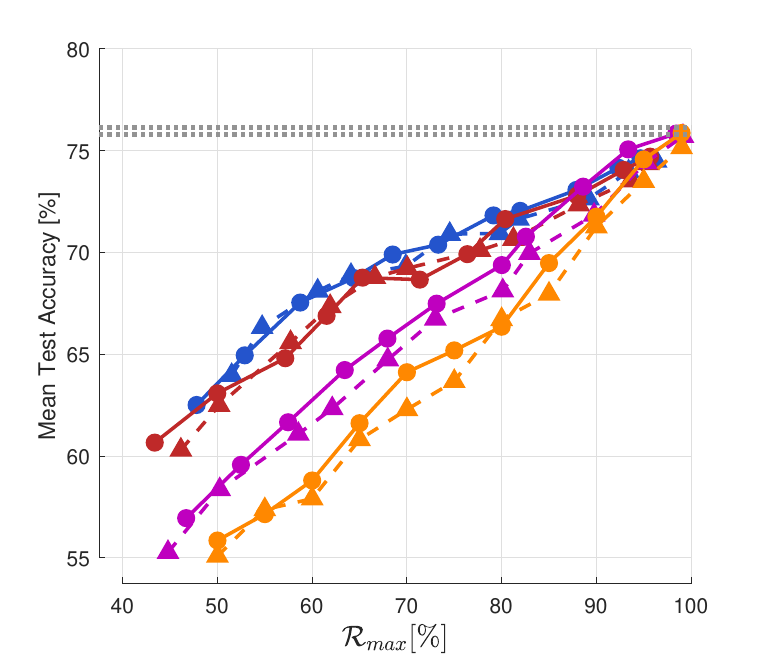}  \caption{$M=4$ nodes}
  \label{fig:sub1}
\end{subfigure}%
\begin{subfigure}{.32\textwidth}
  \centering
    \includegraphics[trim={1cm 0 0cm 0},width=\textwidth]{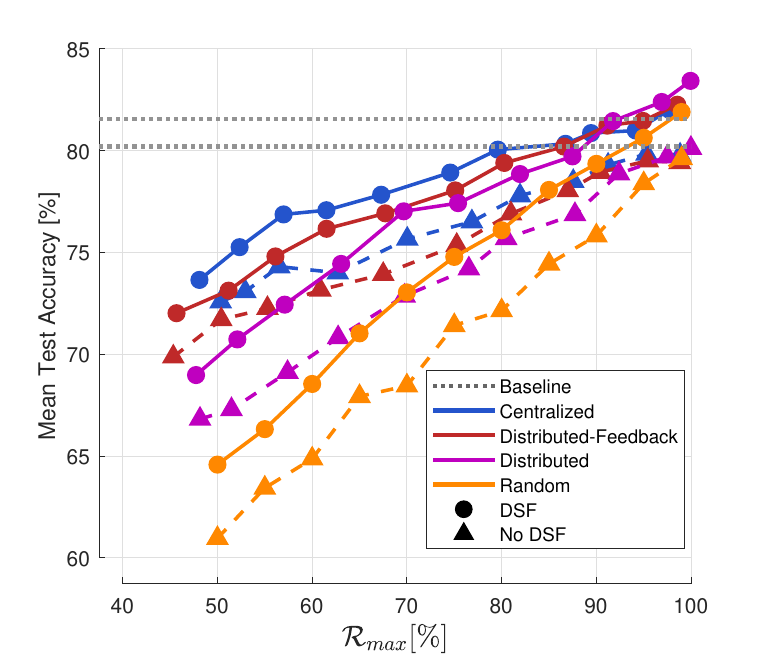}  \caption{$M=8$ nodes}
  \label{fig:sub1}
  \end{subfigure}
  \begin{subfigure}{.32\textwidth}
  \centering
    \includegraphics[trim={1cm 0 0cm 0},width=\textwidth]{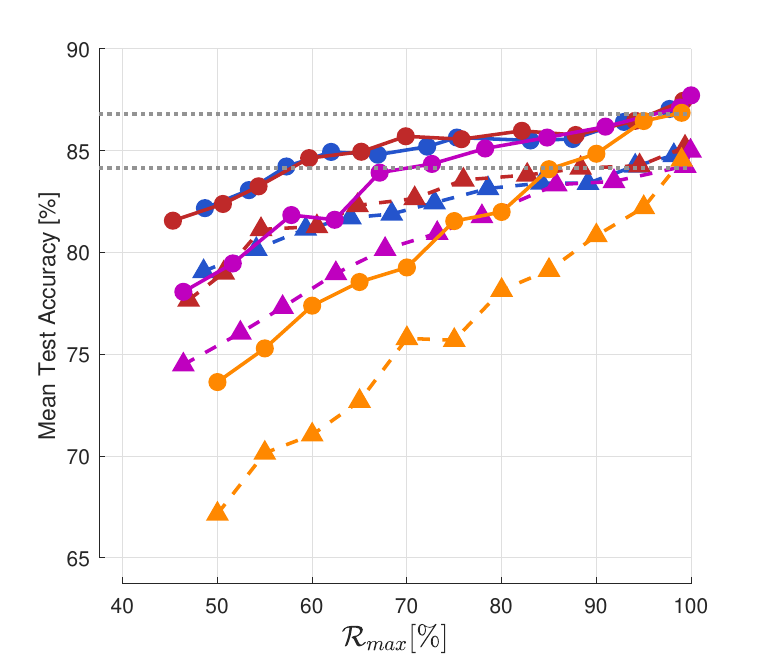}  \caption{$M=16$ nodes}
  \label{fig:sub1}
  \end{subfigure}
\caption{Rate-accuracy trade-off for the proposed dynamic channel selection method for networks of 4,8 and 16 nodes, with and without inclusion of the DSF module. Baseline performance indicates accuracy without dynamic selection involved, i.e. each node transmits at a rate $\mathcal{R}$ of 100\%. While DSF is not necessary for the 4-node network, it becomes more important as the size of the network increases, delivering a consistent performance gain across all settings.}
\label{fig:DSFperformance}
\end{figure*}

\begin{figure}[htbp]
    \centering
    \includegraphics[trim={1cm 0 0cm 0},width=0.4\textwidth]{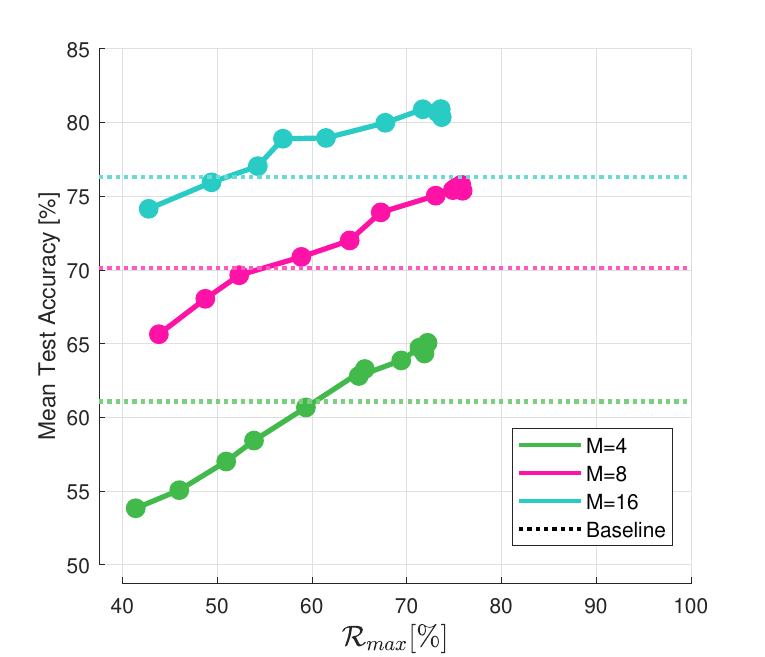}
    \caption{Rate-accuracy trade-off for the proposed dynamic channel selection method in the distributed-feedback setting for networks of 4, 8 and 16 nodes in a simulation of a noisy environment where each channel has a 25\% probability to be replaced by Gaussian noise. When the model is trained with a target rate above this threshold, it automatically rejects the transmission of the noisy and the resulting rate is capped at 75\%. Rejecting these noisy channels yields higher accuracies than the baseline network which also accepts the noisy channels as input, demonstrating the network is now more robust against this noise.}
    \label{fig: noisy}
\end{figure}

Next, we analyze the importance of the presence of the DSF module by comparing it with the networks where it has not been added. The results are illustrated in Fig. \ref{fig:DSFperformance}. For the 4-node network, it can be observed that the inclusion of DSF only results in a small improvement for the random selection network and no improvement at all on the dynamic selection network. As mentioned in section \ref{section: Section2d}, the main purpose of the DSF module is to increase the capability of the classifier to cope with the different channel subsets it is presented with. In this small, 4-node network, the amount of channel subsets $2^M$ is still manageable. Furthermore, in contrast to the random selection, the dynamic selection will not randomly sample from \textit{all} possible channel subsets, but will focus its sampling on a more select number of these. For instance, it will almost always make sure that both a node from the left and right hemisphere is included, since the difference between both is highly informative for motor execution. This explains why the inclusion of DSF is slightly more necessary - and thus beneficial - in the case of random selection than in the case of dynamic selection. When moving to 8- and 16-node networks, the amount of subsets quickly grow and the performance gains afforded by DSF become more and more salient, with these gains once again being slightly higher for the random selection than for the dynamic  selection. 

\subsection{Impact of noisy environments}
\label{section: Section4c}

Up until now, we have discussed situations where we perform a trade-off between the amount of channels we reject and the accuracy of the classifier. In some cases however, working with only a subset of the channels can actually be beneficial for the accuracy as well. In environments where sudden noise bursts can occur, these unexpected inputs, even when limited to a single channel, can heavily disturb the activations of the entire neural network network and lead to misclassifications. To make it easier for the network weights to be robust against these noise bursts, it can be beneficial to detect when these happen and zero the corresponding input instead. To test this hypothesis, we repeated our previous experiments with the dynamic selection method, but in this case, each channel of each input window had a 25\% chance to be replaced by Gaussian noise with zero mean and standard deviation uniformly sampled between 0 and 3 instead, leaving no more relevant information on this channel. Fig. \ref{fig: noisy} compares the performance of the distributed-feedback dynamic selection with a baseline network, which directly takes the perturbed data as input. The noise is added during both training and testing to enable a fair comparison, i.e., the network without dynamic channel selection can in principle learn how to cope with these noise bursts. Firstly, it can be observed that the dynamic selection never transmits more information than absolutely necessary: only 75\% of the channels actually contain information, so the resulting rate is automatically capped around 75\%. Secondly, the automatic rejection of noisy channels does indeed lead to an increased accuracy compared to the baseline accepting this noise as input. A probable reason is that it will be easier for the classifier to find weights that process normal inputs normally and minimize the impact on the activations of disturbances when these disturbances are zero inputs rather than noise bursts.

\section{Conclusion and future outlook}

We have proposed a dynamic channel or sensor selection method in order to reduce the communication cost and improve the battery lifetime of WSNs. For each input window, the method selects the optimal subset of sensors to be used by a neural network classifier, while optimizing a trade-off between the amount of channels selected and the accuracy of the given task. The dynamic selection and the classifier are jointly trained in an end-to-end way through backpropagation. The dynamic selection module consists of three major parts: a channel scoring function assigning a relevance score to each channel, a binary Gumbel-Softmax trick converting these scores to discrete decisions and the dynamic spatial filtering module of Banville et al. \cite{banville2022robust} to make the classifier more robust against the resulting absence of channels. A crucial aspect of this dynamic selection is that it can computed in a \textit{distributed} way, requiring minimal communication overhead between the nodes.
\newline

We have demonstrated the use of this method to perform a trade-off between the transmission rate of the nodes in an emulated wireless EEG sensor network and the accuracy of a motor execution task. Additionally, we have presented a use case where the dynamic selection can even improve the accuracy of the model, by automatically rejecting inputs that might harm performance, such as heavy bursts of noise. Though we have focused on the application use case of wireless EEG sensor networks, our methodology is generic and can be applied to sensor networks with any kind of modalities. In future work, we will explore applications of this method in other distributed platforms than WESNs.
\label{section: Section5}

%\section*{Acknowledgements}

\bibliographystyle{IEEEtran}
%\bibliography{./bibliography/IEEEabrv,./bibliography/IEEEexample}
% \bibliography{mybib.bib}

\onecolumn

\appendices

\section{MSFBCNN architecture}

\begin{table*}[!h]
\begin{tabular}{llllllll}
\hline
\textbf{Layer} & \textbf{\# Filters} & \textbf{Kernel} & \textbf{Stride} & \textbf{\# Params} & \textbf{Output} & \textbf{Activation} & \textbf{Padding} \\ \hline
Input          &                     &                 &                 &                    & (C,T)           &                     &                  \\
Reshape        &                     &                 &                 &                    & $(1,T,C)$       &                     &                  \\
Timeconv1      & $F_T$               & $(64,1)$        & $(1,1)$         & $64F_T$            & $(F_T,T,C)$     & Linear              & Same             \\
Timeconv2      & $F_T$               & $(40,1)$        & $(1,1)$         & $40F_T$            & $(F_T,T,C)$     & Linear              & Same             \\
Timeconv3      & $F_T$               & $(26,1)$        & $(1,1)$         & $26F_T$            & $(F_T,T,C)$     & Linear              & Same             \\
Timeconv4      & $F_T$               & $(16,1)$        & $(1,1)$         & $16F_T$            & $(F_T,T,C)$     & Linear              & Same             \\
Concatenate    &                     &                 &                 &                    & $(4F_T,T,C)$    &                     &                  \\
BatchNorm      &                     &                 &                 & $2F_T$             & $(4F_T,T,C)$    &                     &                  \\
Spatialconv    & $F_S$               & $(1,C)$         & $(1,1)$         & $4CF_TF_S$         & $(F_S,T,1)$     & Linear              & Valid            \\
BatchNorm      &                     &                 &                 & $2F_S$             & $(F_S,T,1)$     &                     &                  \\
Non-linear     &                     &                 &                 &                    & $(F_S,T,1)$     & Square              &                  \\
AveragePool    &                     & $(75,1)$        & $(15,1)$        &                    & $(F_S,T/15,1)$  &                     & Valid            \\
Non-linear     &                     &                 &                 &                    & $(F_S,T/15,1)$  & Log                 &                  \\
Dropout        &                     &                 &                 &                    & $(F_S,T/15,1)$  &                     &                  \\
Dense          & $N_C$               & $(T/15,1)$      & $(1,1)$         & $F_S(T/15)N_C$     & $N_C$           & Linear              & Valid            \\ \hline
\end{tabular}
\caption{Architecture of the MSFBCNN used for motor execution classification. This table is cited from \cite{wu2019parallel}.}
\end{table*}

\end{document}